# Multi-task Learning for Monocular Depth and Defocus Estimations with Real Images

Renzhi He, *Student Member, IEEE*, Hualin Hong, Boya Fu, Fei Liu

**Abstract**—Monocular depth estimation and defocus estimation are two fundamental tasks in computer vision. Most existing methods treat depth estimation and defocus estimation as two separate tasks, ignoring the strong connection between them. In this work, we propose a multi-task learning network consisting of an encoder with two decoders to estimate the depth and defocus map from a single focused image. Through the multi-task network, the depth estimation facilitates the defocus estimation to get better results in the weak texture region and the defocus estimation facilitates the depth estimation by the strong physical connection between the two maps. We set up a dataset (named ALL-in-3D dataset) which is the first all-real image dataset consisting of 100K sets of all-in-focus images, focused images with focus depth, depth maps, and defocus maps. It enables the network to learn features and solid physical connections between the depth and real defocus images. Experiments demonstrate that the network learns more solid features from the real focused images than the synthetic focused images. Benefiting from this multi-task structure where different tasks facilitate each other, our depth and defocus estimations achieve significantly better performance than other state-of-art algorithms. The code and dataset will be publicly available at https://github.com/cubhe/MDDNet.

**Index Terms**—depth estimation, defocus estimation, All-in-3D dataset, multi-task learning, point spread function

---

## 1 INTRODUCTION

Depth and defocus are two useful physical information for the computer to perceive the environments. Many depth estimation [4], [5], [6], [18] and defocus estimation [7], [8], [9], [10] methods are proposed to solve the problems. However, most methods treat the depth and defocus estimations as two separate tasks. From the perspective of computational photography, the depth map and defocus map are encoded into the focused images by the point spread function (PSF) [1], [2], [3]. Theoretically, the depth and defocus map can be decoded from a focused image.

Monocular depth estimation (MDE) is considered to be an ill problem, but deep learning-based methods show impressive results. The previous methods are considered to learn geometric features or perspective relationships in the image [4], [5]. Many methods [3], [12], [27] use other physical information to assist the process of depth estimation where the amount of defocus is a strong cue for an object's depth which provides rich physical information.

Depth from defocus (DFD) methods [6], [12], [17] are proposed to obtain depth maps from focused images. DFD is supposed to acquire more depth information from the defocus cue. [3] is based on phase-coded aperture. [6] uses the physical information between the depth and defocus map. The focused images with different depths of field are used in [17], [43]. The predicted depth map is used in generating the focused images and computing the loss on the focused images [18].

Defocus map estimation (DME) is also of great significance for image processing. DME is applied to obtain advanced image features and be used in many scenarios, such as deblurring [7], [8], image segmentation [9], etc. Some methods achieve defocus estimation by using particularly designed hardware, for example, light field cameras [10], custom lenses [11], and special apertures [12]. We discuss the focused images taken by conventional cameras. Some existing methods [13], [14] first estimate the defocus blur of the edges in the image and then spread it to the whole image. Recently deep learning methods are used to estimate defocus maps. Lee *et al.* [14] propose a defocus map estimation method using domain adaptation to learn defocus blur from multi-scale features, Tang *et al.* [16] find a more accurate defocus map by fusing and refining discriminative multi-scale deep features.

Intuitively, a focused image is supposed to contain more information about depth than a clear image. This is because the depth is implicitly encoded in the focused image. Using the focused image, we can estimate both the depth map and the defocus map. There is also a strong physical connection between the two maps [2], [3].

However, most methods treat defocus and depth estimation as two separate tasks. Without depth, it is hard to estimate the defocus of non-textured areas. Without properly handling the defocus cues, defocus may be treated as noise for depth estimation [5], but this noise contains strong physical information [6], [12]. We aim to make the neural network understand this ambiguous information to facilitate depth and defocus estimations.

Thus, based on the depth estimation network [5], we introduce a defocus map estimation subnet to form a multi-task learning structure [19], [20], [21]. Our network uses an encoder based on vision transformer [22] with two decoders based on selective features fusion (SFF) structure [5]. Because of the high correlation between the defo-cus map and the depth map, this multi-task structure learns transferable features during the training process, which facilitates the network to learn more valid features and the physical relationship between the two tasks. In addition, we derive a physical consistency loss function [6] between the depth map and the defocus map based on PSF.

---

*Renzhi He, Hualing Hong, Boya Fu, Fei Liu are with State Key Laboratory of Mechanical Transmission and the School of Mechanical and Vehicle Engineering, Chongqing University, E-mail: cubhe@foxmail.com; 907968663@qq.com; boya_f@163.com; fei_liu@cqu.edu.cn.*

Additionally, the new network structure imposes requirements for a dataset that contains the focused images, depth maps, and defocus maps. However, there is no dataset that contains these data. Most existing depth-related datasets, such as NYU [24], Make3d [4], and KITTI [25], consisting of only clear images and depth maps, do not contain defocus maps and focus depth. For datasets containing defocused images, such as CUHK [26], Lightfield [27], these datasets do not contain depth and focus depth.

Many DFD methods [6], [11], [18] assume a focus depth and then use this focus depth and the PSF to synthesize focused images for the datasets that only contain the depth maps and clear images. But the PSF models they used do not accurately synthesize realistic focused images, where only defocus is taken into account, diffraction, aberrations, etc. are not taken into account. For focused images generated from datasets such as NYU-V2 and KITTI, which may not have realistic counterparts lenses with large field of view (FOV) with small depth of field (DOF) in the real world.

Thus, we setup a high-precision ground-truth dataset called the **All-in-3D** dataset, consisting of **all-in-focus images, focused images with focus depth, depth maps, and defocus maps**.

Our experiments are conducted with real focused images and highly accurate and dense depth maps. Our goal is to enable the network to learn the real defocus features through real data to achieve highly accurate depth and defocus estimations. Our contributions are as follows:

- We propose a multi-task learning network for depth and defocus estimation, which efficiently unites the depth and defocus map estimation.
- We set up the ALL-in-3D dataset which is the first all-real image dataset consisting of all-in-focus images, focused images with focus depth, depth maps, and defocus maps. The ALL-in-3D dataset is high resolution and precision and contains 100K sets of images.
- We explore the implied defocus and the depth information in the focused images with the All-in-3D dataset and the new network structure. The experiment results demonstrate that the depth and defocus map promote each other by the multi-task architecture.

## 2 RELATED WORK

### 2.1 Depth Estimation

Monocular depth estimation is a challenging area. There are many methods proposed for monocular depth estimation. Saxena *et al*. [28] used a discriminatively-trained Markov Random Field to incorporate multiscale local- and global-image features. Then, with the development of CNNs, Li *et al*. [29] proposed CRF-based end-to-end networks. Recently, Eigen *et al*. [30] used multi-scaled depth estimation. This method was inherited by many methods because it was well integrated with the encoder-decoder structure [31], [32]. More recently, thanks to the development of the vision transformer (VIT) [7], this method was applied to depth estimation [34], [35], [36], [5], e.g., Xie *et al*. proposed a VIT-based method and enlarged the size of receptive field [36]; based on this, Doyeon *et al*. developed a hierarchical transformer encoder and used skip connection to fuse features of different scales [5].

### 2.2 Depth from Defocus

Some methods used the defocus cue to facilitate depth estimation. Zhang *et al*. [6] proposed two networks to estimate the depth map and the defocus map separately and evaluated the depth and defocus maps using a physical consistency-based loss function. However, this physical consistency loss requires an accurate PSF. Shir Gur *et al*. [18] proposed a new neural network structure where the network estimates a depth map from a clear image. In the training phase, instead of calculating the loss directly on depth, a focused image was generated on the depth map and then computed the loss based on the focused image. Ikoma *et al*. [3] proposed a depth from defocus method, where they first made a learned phase-coded aperture and generated the focused image by a corresponding phase-coded method. By this encoding method, the depth information implied in the focused image was better retrieved. Song *et al*. [17], [43] proposed a depth estimation network based on two focused images. Lu *et al*. [44] proposed a self-supervised depth estimation method.

### 2.3 Defocus Estimation

The defocus map is usually considered as consisting of the degree of defocus blur per pixel of the focused image or the CoC size per pixel of the focused image [38]. For the former approach, many algorithms were dedicated to estimating the amount of edge blur accurately, for example, Karaali *et al*. proposed a CNN-based feature learning method [37], Junyong *et al*. proposed an algorithm based on domain adaptation [38]. Shi *et al*. developed a multi-scale solution for blur perception and build a CUHK dataset [26]. There were also shape from focus-related algorithms that were proposed for defocus analysis based on image quality [39], [40]. For the second approach, which was often used in the field of computational photography, the CoC size calculation formula was mentioned in [26], [6], [18], [41]. However, many parameters in the formula, such as focus depth and aperture parameters, were difficult to measure [42].

### 2.4 Related Dataset

The existing datasets related to "all-in-focus image, defocus image and focus depth, depth map, defocus map" can be roughly divided into two categories. The first category is the "clear image, depth image" dataset. For example, NYU-v2 [24], Make3D [4], KITTI [25], FlyingThings3D [45]. The second category is "focused image, defocus map" datasets, such as CUHK [26], DUT [46]. No datasets obtain "all-in-focus image, defocus image, depth map, defocus map, focus depth" for the same viewpoint in the same scene, mainly because of the hardware limitation. Thus, many people have to synthetic focused images based on the first category da-tasets. As listed in Table 1, there is no higher quality dataset available given the current information gathering capabilities of the authors. Our dataset in the last row of Table 1 contains the highest resolution of the images and depth maps and contains ground-truth data rather than synthetic. The dataset will be introduced in detail in Sect. 4.

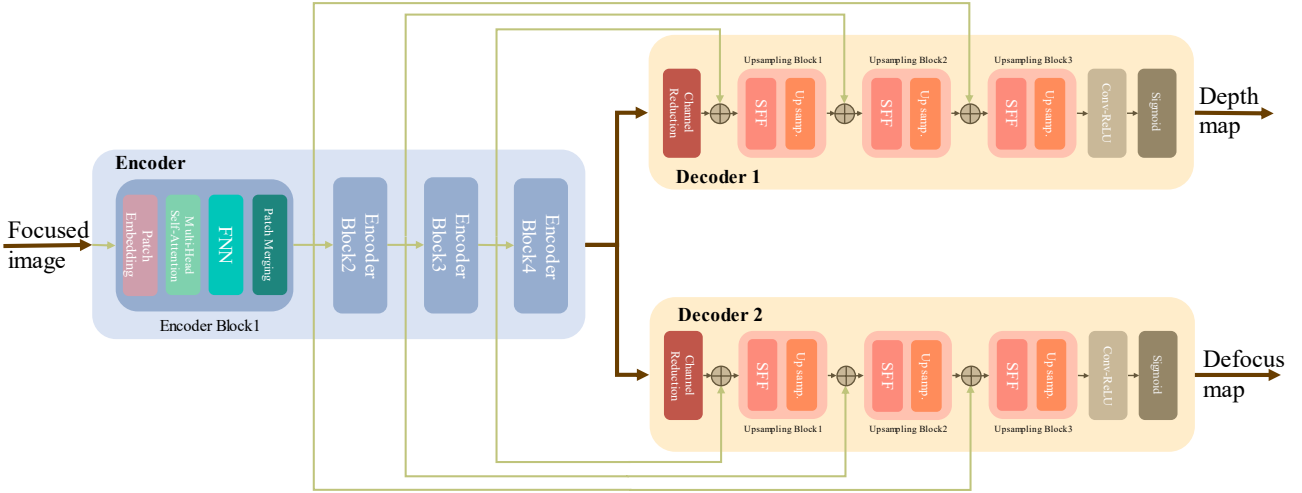

Fig. 1. MDDNet structure overview. Our network consists of one encoder and two decoders, which are connected by skip-connections. The encoder consists of four vision transformer modules and the decoders consist of three SFF modules.

decoder architecture, which is used in depth estimation and semantic segmentation. More specifically, we encode the image with 4 encoder blocks and generate the depth map and the defocus map using two decoders, respectively. The GLPNet is adopted as our backbone network which achieves state-of-the-art performance over the NYU Depth V2. We

TABLE 1
COMPARISON OF EXISTING DATASETS WITH OUR DATASET.

| Name | Number | Resolution[1] | Clear image[2] | Depth map | Focused image | Defocus map |
|---|---|---|---|---|---|---|
| NYU-v2 [24] | 120K | 640×480 | Kinect | Kinect | \ | \ |
| Make3D [4] | 534 | 460×345 | Camera | Laser | \ | \ |
| KITTI [25] | 93K | 370×1226 | Stereo camera | LiDAR | \ | \ |
| FlyingThings3D [45] | 30K | 960×540 | Synthetic | Synthetic | \ | \ |
| CUHK [26] | 1K | 640×427 | \ | \ | Camera | Human labeled |
| DUT [46] | 1.1K | 320×320 | \ | \ | Synthetic | Binary |
| CTCUH [16] | 150 | 480×320 | \ | \ | Camera | Binary |
| SYNDOF [38] | 8231 | 640×480 | \ | \ | Synthetic using PSF | Synthetic CoC map |
| Song *et al.* [17] | 299 | 640×480 | \ | LiDAR | Camera | \ |
| DSLR [47] | 110 | 654×432 | \ | 3D sensor | Camera | \ |
| Ikoma *et al.* [3] | 30K | 960×540 | FlyingThings3D | FlyingThings3D | Synthetic | \ |
| D²Dataset [6] | 1449 | 640×480 | NYU-v2 | NYU-v2 | Synthetic using PSF | Synthetic CoC map |
| **Our dataset (All-in-3D)** | 100K | 2452×2056 | Camera (All-in-focus) | Camera with projector | Camera | **Ground-truth** CoC map |

[1] *The resolution is the lower one between the image and the depth map. The RGB images usually need to down-sample to be aligned with the depth map.*
[2] *clear image means an image with a large DOF; all-in-focus image means the diffuse circle size at each point is theoretically 0.*

## 3 PROPOSED METHOD

### 3.1 Overview

The goal of our model is to predict a depth map and a defocus map from a focused image. To this specialized problem, a **M**ulti-task learning network structure is adopt to estimate the **D**epth and **D**efocus map called **MDDNet**.

As shown in Fig. 1, this network is based on encoder-detail the proposed architecture in the following subsections. This multi-task architecture is supposed to learn the transferable features between depth and defocus map to make the neural network more stable and efficient in processing each task.

### 3.2 The Circle of Confusion (CoC)

First, a formula for the size of the CoC is derives to calculate the ground-truth defocus map and the physical consistency loss, as detailed in the Sect. 3.4.

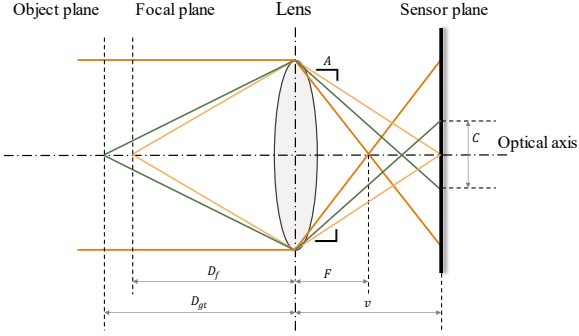

Fig. 2. A simple model of lens. The yellow line represents the focus point. The green line represents the defocus point with a CoC.

Considering the symmetrical lens as illustrated in Fig. 2, the focal length $F$, the distance from the optical lens to sensor plane $v$ and the focus depth $D_f$ satisfy the following equation:

$$\frac{1}{D_f} + \frac{1}{v} = \frac{1}{F}. \quad (1)$$

As seen in Fig. 2, the diameter of the CoC can be written as:

$$C_{pix} = \frac{C_{mm}}{\rho \cdot s} = \frac{a}{\rho \cdot s} \frac{|D_{gt} - D_f|}{D_{gt}} \frac{F}{D_f - F} \quad (2)$$

where $D_{gt}$ is the distance between an object and the lens, and $a = F/N$ where $N$ is the f-number of the camera. While CoC is usually measured in millimeters ($C_{mm}$), we transform its size to pixels by considering a camera pixel-size $\rho$, and a camera output scale $s$.

### 3.3 Multi-task Network for Depth and Defocus Estimation (MDDNet)

In order to combine depth estimation with defocus estimation, we first investigate the relationship between depth and defocus in (2), which are two different physical quantities but share many common features. Based on the physical properties of the depth and defocus maps, the multitask structure network is used to estimate the two maps, which is a learning paradigm in machine learning. It is expected to learn the common features of these two physical quantities to improve the generalization performance of the two tasks.

Specifically, one encoder and two decoders are used to achieve this multi-task structure, where the two decoders share the same encoder and separately estimate the depth map and defocus map. Between the encoder and each decoder, a skip connection is used to make the encoder recover extracted features to the maps.

In the encoder, a multi-scale structure is used to encode the images to ensure that the network learns the features at different scales. More specifically, we use 4 serially connected vision transformer modules to form the encoder to generate features of 1/4, 1/8, 1/16, and 1/32 of the original image size. These features are fed to the decoder via skip connection. In the decoder, we up-sample the features obtained from the encoder to get the depth map and defocus map of the original image size. The SFF module is used to fuse the local and global features during each up-sampling process. Finally, the depth map and the defocus map are generated with two decoders respectively.

### 3.4 Training Loss

Since our network has two outputs, it is necessary to calculate the distance between predicted depth and ground-truth depth, and the distance between the predicted defocus map and ground-truth defocus map.

For depth estimation, scale-invariant log scale loss proposed by Eigen *et al*. [30]. has been widely used. For defocus map evaluation, many people use SSIM and MSE [54]. We use structural similarity loss and scale-invariant log scale loss.

$$\mathcal{L}_{depth} = \frac{1}{N} \sum \left[ \lambda_1 \frac{1 - SSIM(\widehat{D}, D)}{2} + \lambda_2 \left( d_{dp}^2 - \frac{\sum d_{dp}^2}{2N} \right) \right], \quad (3)$$

$$\mathcal{L}_{defocus} = \frac{1}{N} \sum \left[ \lambda_3 \frac{1 - SSIM(\hat{J}, J)}{2} + \lambda_4 \left( d_{df}^2 - \frac{\sum d_{df}^2}{2N} \right) \right] \quad (4)$$

where $\widehat{D}$ and $D$ is the predicted depth and the actual depth, $\hat{J}$ and $J$ is the predicted defocus map and the actual defocus map. $d_{dp} = \log D - \log \widehat{D}$, $d_{df} = \log J - \log \hat{J}$.

For the two outputs of the network, physical consistency loss is used. This loss can decrease the network overfitting, and improve the results of both networks.

$$\mathcal{L}_{pc} = \lambda_5 \frac{1}{N} \left\| \hat{J} - A \frac{|\widehat{D} - D_f|}{\widehat{D}} \frac{F}{D_f - F} \right\|_2 \quad (5)$$

where $A := a/(\rho \cdot s)$.

So the loss function of the whole network can be written as:

$$\mathcal{L} = \mathcal{L}_{depth} + \mathcal{L}_{defocus} + \mathcal{L}_{pc}. \quad (6)$$

## 4 THE ALL-IN-3D DATASET

In Table 1 the existing datasets are discussed, which cannot meet the demand of high precision focused image, depth map, defocus map in the same view. Thus, we build a dataset containing **all-in**-focus image, focus**d** image with focus depth, **d**epth map and **d**efocus map, called the **All-in-3D** dataset. In addition, we derive the synthetic focused image generation algorithm and generate focused images on our dataset for the subsequent comparison experiments. Compared to the existing datasets, listed in Table 1, the main advantages of our dataset are as follows:

A. The high-resolution RGB images and depth maps with the size of 2452 × 2056 are provided.
B. The depth maps in our dataset do not require extra alignment or interpolation with the RGB images, since the depth is solved by pixel [53]. Compared to the depth maps obtained by laser, the depth maps in our dataset are dense and have the same FOV as the RGB images.
C. It provides pixel-level annotated defocus maps where the defocus is calculated by the CoC size (2). See Table 1, [16], [46] provide the binary defocus map labeled by subjective perception or clearness analysis, and [38], [6] use the simulated focus depth to calculate the CoC size of the defocus maps.
D. More general. The dataset can be applied in SFF/SFDF and optical deblur domain.

### 4.1 Data Collection for Our Dataset

The All-in-3D dataset contains 500 scenes. Each scene is focused 200 times to generate 200 focused images with different focus depths, where the focused images share the same depth map and all-in-focus image. Our dataset contains 100K (500×200) sets of data. A set of data consists of an all-in-focus image, a focused image with focus depth, a depth map and a defocus map.

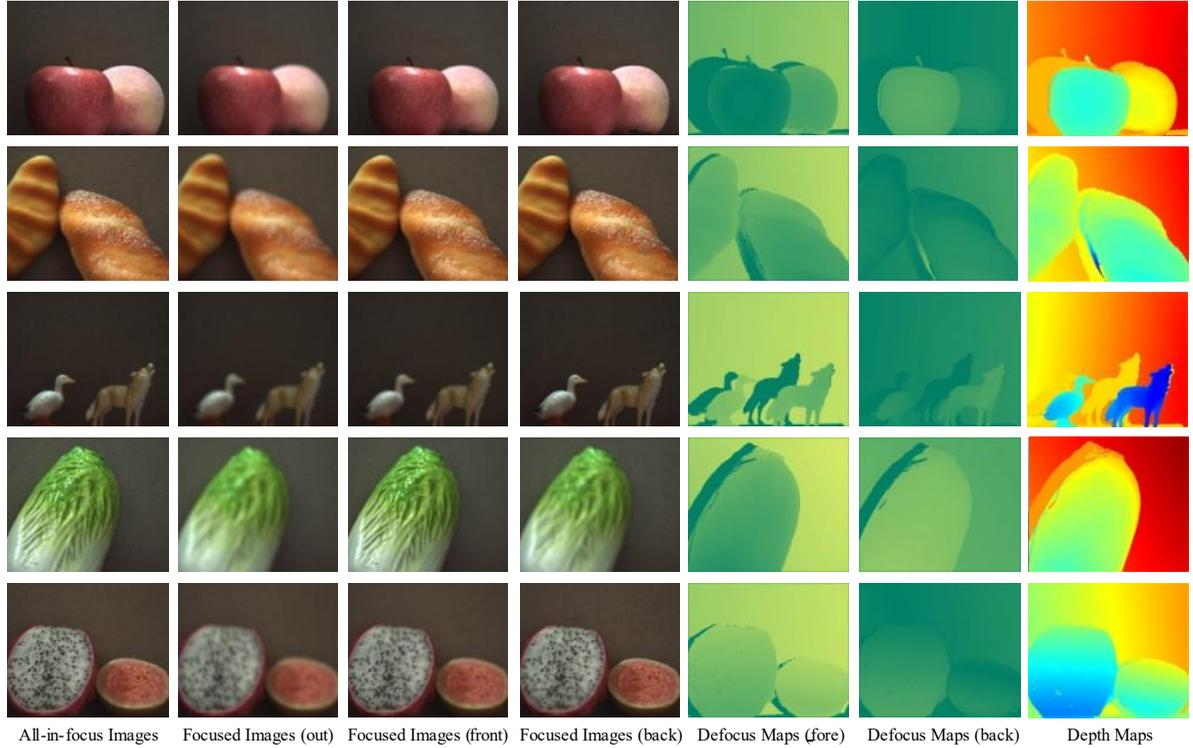

Fig. 3. Examples of the All-in-3D dataset. The first column is the all-in-focus images. The second column is the focused images that every point is out-of-focus. The third and fourth columns are the focused images that focused on the front and the back. The fifth and sixth columns are the defocus maps corresponding to columns the third and fourth columns. The seventh column is the corresponding depth maps.

**Focused image** is obtained by direct camera photography. The resolution of the image is 2452 × 2056. The maximum aperture of our lens is #1.4 and the focal length is 50mm when shooting focused images. At the maximum aperture, focused images are obtained with a small DOF. For each scene, 200 focused images are taken with different focus depths.

**All-in-focus image** is obtained by two methods. It can be obtained by compositing the focus images. Similar to the SFF algorithm [40], a focus stack is generated from 200 images, from which the clearest sequence of each pixel position is selected to form an all-in-focus image. The algorithm is shown in Table 2. Another method is to set the lens aperture to the minimum #16 and use the camera to take an image of a large DOF.

**Depth map** is obtained by the fringe projection profilometry method (FPP) [53] based on the structured light system. Using the FPP method, the depth map is solved by pixel. Thus, the resolution and the FOV of the depth map are the same as the focused images and all-in-focus images. The depth map does not require extra calibration or interpolation to be aligned with the focused images.

**Depth map refinement $D_{gt}$.** However, the projector and camera in the structured light system are a binocular system, with parallax. As shown in Fig. 4 (a)(c), the shadowed or reflective parts cannot be solved by the FPP method. Firstly, the area that cannot be solved by structured light to generate a mask is identified, as shown in Fig. 4 (b). The unsolvable area is identified and filled by the depth $D_{sf}$ solved by the SFF algorithm, see Table 2. Fig. 4 (c) is the original depth map solved by the FPP method. Fig. 4 (d) is the refined depth map.

**Focus depth $D_f$.** The distance from the camera CCD (or CMOS) to the lens plane is measured by the infrared sensor. The resolution of the infrared sensor is 3μm and it is calibrated in [2] to achieve high accuracy of the distance from the camera CCD to the lens plane. After obtaining the distance $v$, the focused depth $D_f$ can be calculated by (1)

TABLE 2
ALL-IN-FOCUS IMAGE AND DEPTH MAP GENERATION ALGORITHM

| Algorithm 1: All-in-focus image generation |
|---|
| Input: Focused image stack $[I_1\ I_2\ ...\ I_T,] \in R^{H \times W \times C \times T}$, focus depth $fd \in R^T$ |
| Output: Depth map $D_{sf}$, all in focus image $I_c$ |
| 1. Apply Laplace operation to each image to generate Laplace stack $$[\sigma_1\ \sigma_2\ ...\ \sigma_T,] \in R^{H \times W \times T}$$ |
| 2. Search for the index of the maximum value on the Laplace stack $$ids(i,j) = argmax([\sigma_1(i,j), \sigma_2(i,j)\ ...\ \sigma_T(i,j)])$$ |
| 3. Compose an all-in-focus image and depth map $\quad for\ i = 0\ to\ H\ do$ $\quad\quad for\ j = 0\ to\ W\ do$ $\quad\quad\quad I_c(i,j) = I_{ids(i,j)}(i,j)$ $\quad\quad\quad D_{sf}(i,j) = fd[ids(i,j)]$ |

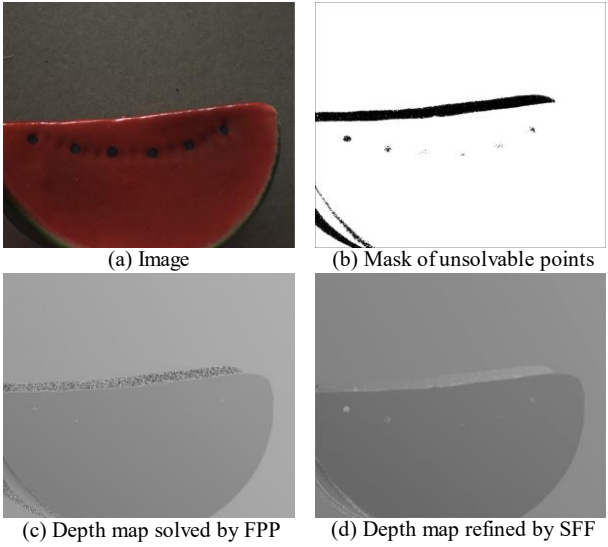

Fig. 4. Process of depth refinement. (a). The RGB image. (b) The mask consisted of unsolvable points. (c) The depth map solved by FPP method. (d) The depth map refined by SFF.

**Defocus map** is calculated by the value the CoC size following (5):

$$J(i,j) = A \frac{|D_{gt}(i,j) - D_f|}{D_{gt}(i,j)} \frac{F}{D_f - F} \quad (7)$$

where A in (7) is 800, solved in [2].

## 4.2 Synthetic Focused Image Generation

In addition, focused images are synthesized based on our dataset for the comparison experiments. First, a PSF function is derived using the Gaussian function [5], [6] and used to compute a CoC for each point of the image and then superimpose all the CoCs.

$$G(x, y, C_{pix}) = \frac{1}{2\pi r^2} exp\left(-\frac{x^2 + y^2}{2C_{pix}^2}\right), \quad (8)$$

$$I_b(x,y) = \sum_{i=x-\frac{k}{2}}^{x+\frac{k}{2}} \sum_{j=y-\frac{k}{2}}^{y+\frac{k}{2}} I_c(i,j) \cdot G\left(x-i, y-j, C_{pix}(D(i,j))\right), \quad (9)$$

where the $I_b$ is the synthetic focused image, $I_c$ is the all-in-focus image in our dataset. Unlike the convolution method [6], [38], this approach is closer to the physical process of the focused images. The whole process is implemented on the GPU, using NVIDIA Cuda toolkits. Fig. 5 shows the real focused images and synthetic focused images.

## 5 EXPERIMENTAL ASSESSMENT

This section discusses the performance of our network in defocus and depth map estimations. First, the experimental setting is introduced. Then the effect of the real and synthetic focused images on the network is discussed. Then our network MDDNet is used to generate the depth and defocus maps and compare our defocus and depth estimations results with other methods. The MDDNet contains two decoders. The network is trained only using the defocus decoder called Defocus subnet and the network is trained only using the depth decoder called Depth subnet. The two subnets are tested in the ablation study to show the effectiveness of each component's contribution.

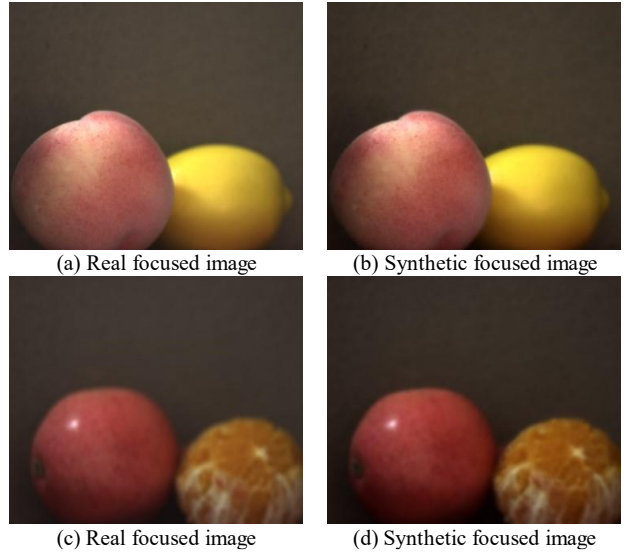

Fig. 5. Comparison of real and synthetic focused images. The real focused images (a) (c) and the synthetic focused images (b) (d) where (a) (b) focus on the front, (c) (d) focus on the back.

## 5.1 Experimental Configuration

The network is implemented with PyTorch framework and trained using Adam [52] optimizer with learning rate $10^{-4}$. For the encoder, the GLP pre-trained weights [5] are used, and for decoders, the initial weights are used. The whole Batch size is 4. All the experiments are conducted with two Nvidia RTX 3090 24G.

## 5.2 Real Images or Synthetic Images?

As shown in Table 1, many methods [3], [6], [38], [46] use synthetic images to train the network. To figure out the effect of synthetic focused images and real focused images on the network. The synthetic focused images are synthesized based on the all-in-focus images with (11). The MDDNet is trained and tested on synthetic images and real focused images.

First, the MDDNet is trained on the synthetic focused images, then tested on the synthetic images and the real images. The results are listed in the first 4 rows in Table. 3. Second, the network is trained on the real focused images and then tested on the synthetic images and the real images. The results are listed in the last 4 rows in Table. 3. Fig. 5 shows the difference between our synthetic focused images and the real focused images.

For the defocus estimation, the network gets the best result when trained on the synthetic focused image tested on the synthetic focused images which is 0.2054 under the $\delta < 1.05$ metric but gets the worst result when the network is trained on the synthetic images and tested on the real focused images which is 0.1499 under the $\delta < 1.05$ metric. When tested on the real images, as shown in Fig. 6 (c)(e), the network trained on the real images gets a better result of 0.1912 which is 5% higher than the network trained on the synthetic images. It means that the network trained on the synthetic focused images cannot transfer to the real focused images and the real focused images make the

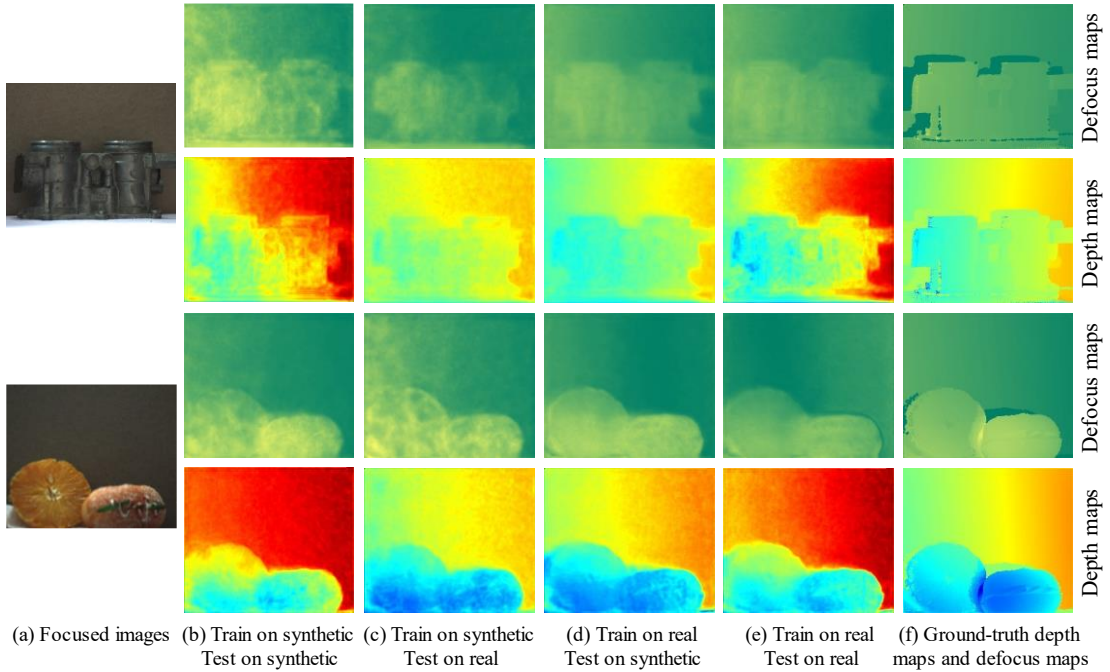

(a) Focused images  (b) Train on synthetic Test on synthetic  (c) Train on synthetic Test on real  (d) Train on real Test on synthetic  (e) Train on real Test on real  (f) Ground-truth depth maps and defocus maps

Fig. 6. Depth and defocus estimation results of the MDDNet with trained and tested on real or synthetic focused images. Results of MDDNet trained on the synthetic focused image then tested on the synthetic focused image (b) and real focused images (c). Results of MDDNet Trained on the real focused image then tested on the synthetic focused images (d) and real focused images (e). The ground-truth defocus maps (f) row 1, 3 and depth maps (f) row 2, 4.

TABLE 3
DEPTH AND DEFOCUS ESTIMATION RESULTS OF MDDNET TRAINED AND TESTED ON SYNTHETIC AND REAL FOCUSED IMAGES.

| Train set | Test set | Output | $\delta < 1.05 \uparrow$* | $\delta < 1.15 \uparrow$ | $\delta < 1.25 \uparrow$ | Abs Rel↓ | Sq Rel↓ | RMSE↓ | RMSE log↓ | Log 10↓ |
|---|---|---|---|---|---|---|---|---|---|---|
| Synthetic | Synthetic | Depth | 0.8455 | 0.9988 | 1.0000 | 0.0273 | 0.0012 | 0.0301 | 0.0320 | 0.0120 |
| | | Defocus | 0.2054 | 0.4580 | 0.6078 | 0.4420 | 0.052 | 0.0849 | 0.4028 | 0.1264 |
| | Real | Depth | 0.8427 | 0.9982 | 1.0000 | 0.0280 | 0.0013 | 0.0307 | 0.0325 | 0.0123 |
| | | Defocus | 0.1499 | 0.3403 | 0.4700 | 0.7471 | 0.134 | 0.1184 | 0.5482 | 0.1869 |
| Real | Synthetic | Depth | 0.8578 | 0.9973 | 1.0000 | 0.0239 | 0.0011 | 0.0267 | 0.0291 | 0.0105 |
| | | Defocus | 0.1610 | 0.3779 | 0.5106 | 0.8192 | 0.1719 | 0.115 | 0.5477 | 0.183 |
| | Real | Depth | 0.8850 | 0.9979 | 1.0000 | 0.0213 | 0.0008 | 0.0239 | 0.0263 | 0.0093 |
| | | Defocus | 0.1912 | 0.4195 | 0.5484 | 0.6412 | 0.1104 | 0.0998 | 0.5168 | 0.1716 |

*We use the same metrics [34] for depth and defocus estimations. The up arrow indicates the larger value achieved the better performance is, while the down arrow indicates the smaller, the better.

network learn the real defocus features.

For the depth estimation, the network gets the best result of 0.8850 under the $\delta < 1.05$ metric when trained and tested on the real focused images, which is 5% higher than the trained on synthetic and tested on real images (0.8427). It means the real defocus features learned from the real focused images, better improve the depth estimation than the synthetic focused images, as shown in Fig. 6 (c)(e).

The synthetic focused images face two problems. First, it is difficult to synthesize an accurately focused image. Most Gaussian-based focused image generation methods are linear and only take into account defocus blur, while realistic focused images have longitudinal and lateral chromatic aberrations etc. Second, the parameters used in the defocus PSF are not realistic. Take NYU-v2 for example, the depth map in NYU-v2 contains data up to ten meters, it is difficult to manufacture such a lens that maintains a small DOF and a large FOV at such a distance.

The experiments prove that the synthetic focused images have some limitations compared to the real focused images. The synthetic images are hard to make the network learn the real features of focus and hard to assist the depth estimation. Therefore, our network is trained on the real focused images.

### 5.3 Defocus Estimation

We implement three algorithms, Xu *et al.* [51], DMENet [14], and Zhang *et al.* [6]. Our method estimate the defocus map based on the size of CoC, while other datasets are based on image quality evaluation [26], [46]. Thus, the defocus maps are normalized to the same scale before comparing. To evaluation the defocus maps, some use the F-measure scores [16], [46], or MAE [14], [16], [46].

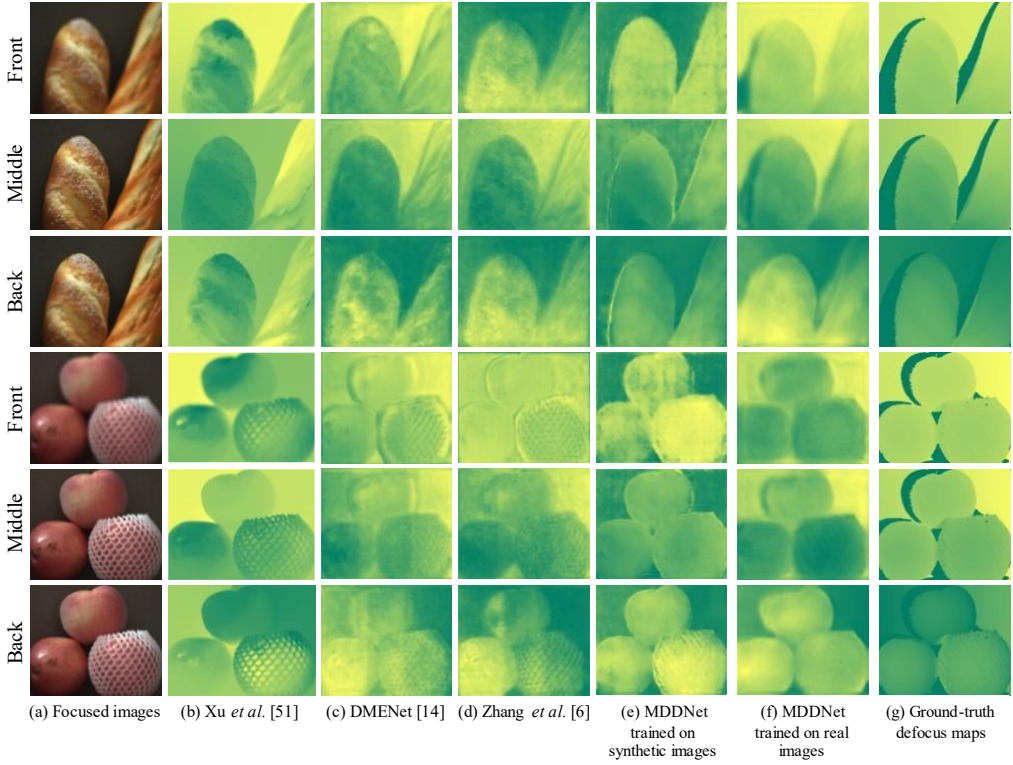

Fig. 7. The defocus estimation results. (a) The focused images focused on the front, middle, and back. (b)-(d) Defocus maps generated by [51], [38], [6] (trained on real images). (e) Defocus maps generated by MDDNet (trained on synthetic images). (f) Defocus maps generated by MDDNet (trained on real images). (G) The ground-truth defocus maps.

TABLE 4
DEFOCUS ESTIMATION RESULTS OF MDDNET AND OTHER METHODS*.

| Methods | $\delta < 1.05 \uparrow$ | $\delta < 1.15 \uparrow$ | $\delta < 1.25 \uparrow$ | Abs Rel $\downarrow$ | Sq Rel $\downarrow$ | RMSE $\downarrow$ | RMSE log $\downarrow$ | Log 10 $\downarrow$ |
|---|---|---|---|---|---|---|---|---|
| Xu *et al.* [51] | 0.0247 | 0.0709 | 0.1139 | 492.3822 | 488.9422 | 0.423 | 2.4087 | 0.6609 |
| DMENet [14] | 0.1309 | 0.3215 | 0.4681 | 0.822 | 0.1521 | 0.1232 | 0.5516 | 0.1832 |
| Zhang *et al.* [6] | 0.1493 | 0.3527 | 0.4853 | 0.7228 | 0.1238 | 0.1000 | 0.5662 | 0.1921 |
| MDDNet (synthetic) | 0.1610 | 0.3779 | 0.5106 | 0.8192 | 0.1719 | 0.115 | 0.5477 | 0.183 |
| MDDNet (real) | 0.1912 | 0.4195 | 0.5484 | 0.6412 | 0.1104 | 0.0998 | 0.5168 | 0.1716 |

*Only the MDDNet (synthetic) is trained on synthetic focused images other methods are trained on real focused images.*

Fig. 7. shows the defocus map estimation for different focus depths on the same scene. It can be seen that the defocus estimation results of our algorithm are visually more sensitive to the defocus and focus areas.
Seeing the apple wrapped in a bubble grid in the bottom half of Fig. 7, where the texture changes drastically, but the depth changes slightly, the defocus estimation methods [51], [14], [6] cannot handle this area properly. These methods pay much attention to the edges. But our method benefits from the depth estimation, it helps the defocus decoder to handle the place where the texture changes drastically, but the depth changes slightly. Similarly, benefiting from the depth estimation, our algorithm handles the weakly textured part properly, like the background. Fig. 8 further demonstrates the accuracy of our algorithm. Seeing the keycaps on the keyboard, our algorithm can distinguish each keycap. Compared to [51], the MDDNet avoids the impact of letters on keycaps.

Table. 4 shows the quantitative analysis of our algo-rithm. For the $\delta < 1.05$ metrics, MDDNet(ours) is 0.1912, while Zhang's method [6] and DMENet [14] method are 0.1493 and 0.1309, respectively. Our method is larger than those methods in [6] (5%) and [14] (6%), which means our defocus estimation algorithm achieves higher accuracy compared to other neural networks and is less affected by the edges.

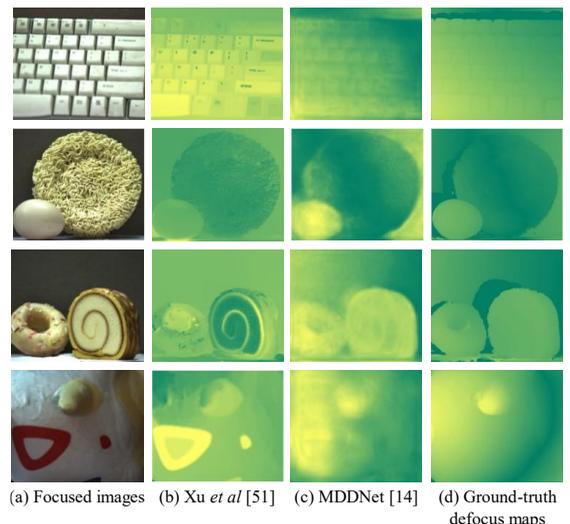

Fig. 8. The defocus estimation results. (a) The focused images. (b) Defocus maps generated by [51]. (c) Defocus maps generated by MDDNet. (d) The ground-truth defocus maps.

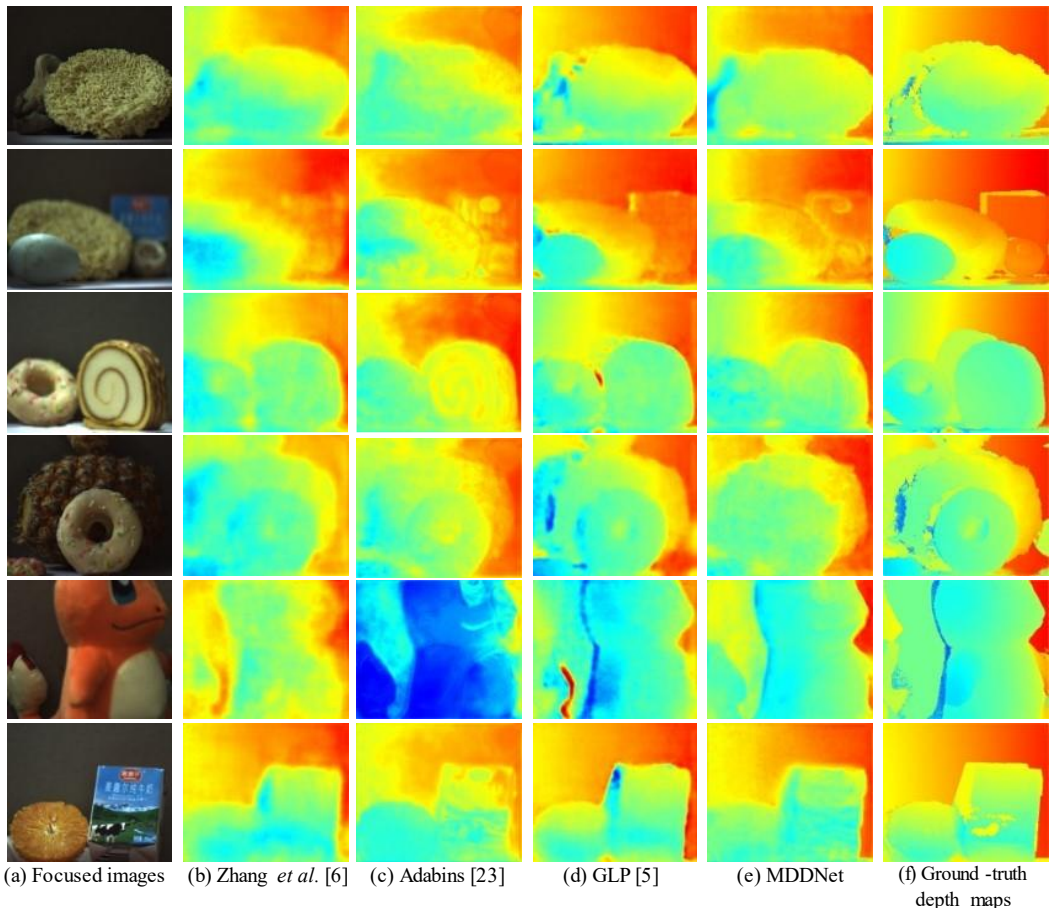

Fig. 9. The depth estimation results. (a) The focused images. (b)-(e) Depth maps generated by [6], [23], [5]. (f) Depth maps generated by MDDNet. (G) The ground-truth depth maps.

TABLE 5
DEPTH ESTIMATION RESULTS OF MDDNET AND OTHER METHODS.

| Methods | $\delta < 1.05 \uparrow$ | $\delta < 1.15 \uparrow$ | $\delta < 1.25 \uparrow$ | Abs Rel $\downarrow$ | Sq Rel $\downarrow$ | RMSE $\downarrow$ | RMSE log $\downarrow$ | Log 10 $\downarrow$ |
|---|---|---|---|---|---|---|---|---|
| Zhang *et al.* [6] | 0.7594 | 0.9794 | 0.9999 | 0.0369 | 0.0022 | 0.0374 | 0.0420 | 0.0156 |
| Adabins [23] | 0.8667 | 0.9912 | 1.0000 | 0.0257 | 0.0013 | 0.0282 | 0.0307 | 0.0112 |
| GLP [5] | 0.8159 | 0.9797 | 0.9999 | 0.0320 | 0.0018 | 0.0325 | 0.0368 | 0.0135 |
| MDDNet (Ours) | 0.8850 | 0.9979 | 1.0000 | 0.0213 | 0.0008 | 0.0239 | 0.0263 | 0.0093 |

### 5.4 Depth Estimation

Several monocular depth estimation algorithms are implemented to compare with our MDDNet, including [6], [23], and [5], as shown in Fig. 9 (b)(c)(d). The result of our algorithm is shown in Fig. 9(e).

As shown in row 5 of Fig. 9, other algorithms have difficulty recovering the tail of the Charmander, while our algorithm can estimate the unsolvable part of the original depth. Table. 5 shows the quantitative analysis of our algorithm with other algorithms. For the $\delta < 1.05$ metrics, our method is 0.8850, while Zhang's [5] method is 0.7594, Adabins is 0.8667, and GLP is 0.8159. Our methods achieves higher accurate than [5] (13%), [23] (2%),and [5] (7%). It means that our network learns more valid accurate depth information from the focused images. The most difference between our depth estimation method and the previous methods is that the defocus estimation is introduced to help the depth estimation.

Then ablation experiments are conducted to further demonstrate the effectiveness of Depth subnet and Defocus subnet and explore how they facilitate each other.

### 5.5 Single Task or Multi-task?

In this subsection, first, the validity of the network structure is verified, where the depth maps are estimated using the encoder with a single depth decoder called Depth subnet, and the defocus map is estimated using the encoder and a single defocus decoder called Defocus subnet. Then the validity of the physical consistency loss is verified.

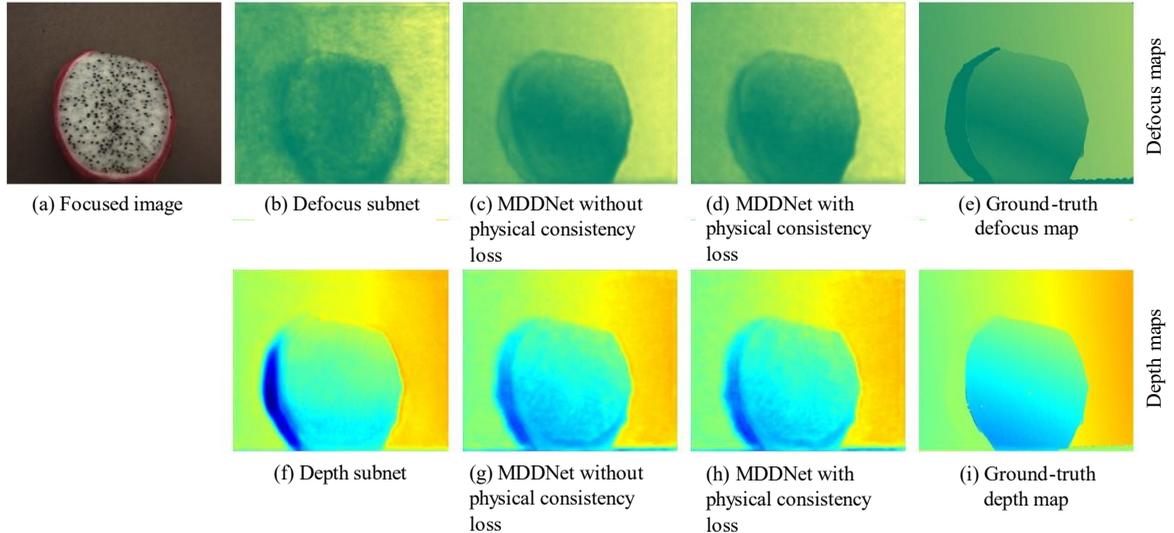

Fig. 10. Ablation study. (a) The focused image. (b)-(e) are the corresponding defocus maps. (f)-(i) are the corresponding depth maps. (b) (f) are generated by two subnets. (c) (g) are generated by MDDNet trained without using the physical consistency loss. (d) (h) are generated by MDDNet trained using the physical consistency loss. (e) (i) are the ground-truth defocus map and depth map.

TABLE 6
ABLATION STUDY. RESULTS OF DEPTH SUBNET, DEFOCUS SUBNET, MDDNET WITHOUT PHYSICAL CONSISTENCY LOSS AND MDDNET WITH PHYSICAL CONSISTENCY LOSS.

| Methods | Output | $\delta < 1.05 \uparrow$ | $\delta < 1.15 \uparrow$ | $\delta < 1.25 \uparrow$ | Abs Rel $\downarrow$ | Sq Rel $\downarrow$ | RMSE $\downarrow$ | RMSE log $\downarrow$ | Log 10 $\downarrow$ |
|---|---|---|---|---|---|---|---|---|---|
| Depth subnet | Depth | 0.8160 | 0.9986 | 1.0000 | 0.0269 | 0.0013 | 0.0296 | 0.0323 | 0.0117 |
| Defocus subnet | Defocus | 0.1610 | 0.3779 | 0.5106 | 0.8192 | 0.1719 | 0.1150 | 0.5477 | 0.1830 |
| MDDNet without physical consistency loss | Depth | 0.8661 | 0.9978 | 1.0000 | 0.0235 | 0.001 | 0.0261 | 0.0284 | 0.0102 |
| | Defocus | 0.1830 | 0.4213 | 0.5542 | 0.7116 | 0.1312 | 0.1022 | 0.4997 | 0.1641 |
| MDDNet | Depth | 0.8850 | 0.9979 | 1.0000 | 0.0213 | 0.0008 | 0.0239 | 0.0263 | 0.0093 |
| | Defocus | 0.1912 | 0.4195 | 0.5484 | 0.6412 | 0.1104 | 0.0998 | 0.5168 | 0.1716 |

**Depth subnet.** Fig. 10 (f) shows the result of depth estimation without Defocus subnet. Compared to Fig. 10 (h), the Defocus subnet improves the depth estimation at the edges and the shadows. Table 6 lists the quantitative analysis of the improvement, the depth estimation results of the Depth subnet and MDDNet under the $\delta < 1.05$ metric are 0.8160 and 0.8850, which means the Defocus subnet improves the accuracy of depth estimation by 5%.

**Defocus subnet.** As discussed in Sect. 5.3, the defocus networks have difficulty in estimating the parts without texture and pay much attention to the edges. This problem is also present in our Defocus subnet. This problem is effectively solved by adding a depth estimation network shown in Fig. 10 (b)(d). For the results of MDDNet, the defocus map is smoother, and the background is clearer than the result of Defocus subnet. The defocus estimation results of the defocus subnet and MDDNet under the $\delta < 1.25$ metric are 0.5106 and 0.5484, which means the Depth subnet improves the accuracy of defocus estimation by 3%.

**Physical consistency loss.** We use this loss function, as shown in Fig. 10 (c)(g)(d)(h), to reduce overfitting, and improve the performance of the network. Table. 6, lists the effect of this loss function on the network, the result of MDDNet trained without physical consistency loss for depth and defocus estimations are 0.8661 and 0.1830 under the $\delta < 1.25$ metric while the MDDNet with the loss are 0.8850 and 0.1912 which means the physical consistency loss improves the depth estimation by 2% and improves the defocus estimation by 1%.

## 6 CONCLUSION

The MDDNet proposed in this paper is demonstrated to be highly competitive, by comparing it with other methods. The multi-task structure encoder can effectively learn the valid information related to depth as well as defocus from the focused image. Moreover, it is more generalized compared to the specially manufactured lenses since we acquire the focused images from the conventional lens.

We setup a dataset consisting of the all-in-focus image, focused image with focus depth, depth map and defocus map that has high precision annotation and containing 100k sets of data. As far as we know, it is the only and the largest dataset that contains focused images with corresponding high accuracy depth maps and precision pixel annotation defocus maps. We believe that this dataset will facilitate the development of defocus/depth estimation and optical deblurring etc. The form of focused images is closer to the way humans perceive objects, and we hope that our dataset and the method proposed in this paper will facilitate the ability of computers or robots in visual perception. For the deficiencies in the dataset, such as partial unsolvable depth, we are working to complement these with neural networks or a multi-view system.

In future work, we will improve the network structure and use more physical information to make the network obtain more accurate depth and defocus maps. We hope to take more insight into the focused images and fully take the advantage of our dataset.

## ACKNOWLEDGMENT

The corresponding author is Fei Liu. This work is supported in part by the Key Technologies Research and Development Program (2018YFB2001400).

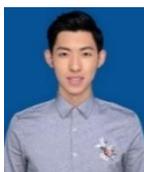

**Renzhi He** received the BS degree in mechatronics engineering form the Chongqing University of Posts and Telecommunications. He is currently studying in Chongqing University for his master's degree. His research interests include computational photography, computer vision, SFF, and structured light. He is a student member of the IEEE.

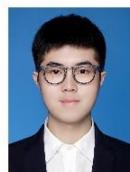

**Hualin Hong** received the engineering degree in Mechatronic Engineering from JiMei University, XiaMen, China, in 2021. He is currently pursuing the master's degree in School of Mechanical and Vehicle Engineering, Chongqing University, Chongqing, China, under the supervision of a.p. F. Liu. His current research interests include depth estimation and defocus estimation in computer vision.

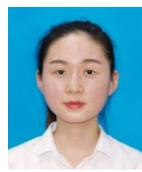

**Boya Fu** received the B.E. degree in machine design manufacture and automation from Southwest Jiaotong University, Chengdu, China, in 2020. She is currently studying as a graduate student in mechanical and electronic engineering in Chongqing University, Chongqing, China.
Her research interests are in computer vision, shape from focus and image processing.

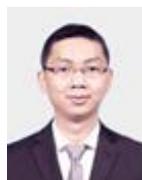

**Fei Liu** received his B.S. degree from Huazhong University of Science and Technology and Ph. D. degree from Tsinghua University, China, respectively. During 2012~2013, he was a visiting scholar in Massachusetts Institute of Technology, USA.
He is an associate professor with the Department of Mechanical Engineering at Chongqing University, China, and a Research Fellow in the State Key Lab of Mechanical Transmission. His research interests include robotics, machine vision, the optical measurement and its application.